\pdfoutput=1

\documentclass[11pt]{article}
\usepackage{graphicx}
\usepackage[final]{acl}

\usepackage{times}
\usepackage{latexsym}
\usepackage{graphicx}
\usepackage{amsmath}
\usepackage{multirow}
\usepackage{todonotes}
\usepackage{booktabs}
\usepackage{float}
\usepackage[T1]{fontenc}

\usepackage[utf8]{inputenc}
\usepackage{amsmath}

\usepackage{microtype}

\usepackage{inconsolata}
\usepackage{amssymb}

\usepackage{array}
\newcolumntype{P}[1]{>{\centering\arraybackslash}p{#1}}

\title{Will LLMs Replace the Encoder-Only Models in \\ Temporal Relation Classification?
}

\author{Gabriel Roccabruna, Massimo 
  Rizzoli, Giuseppe Riccardi\\
  Signals and Interactive Systems Lab\\
  University of Trento, Italy \\
  \texttt{\{gabriel.roccabruna, massimo.rizzoli, giuseppe.riccardi\}@unitn.it}  }

\begin{document}
\maketitle
\begin{abstract}
The automatic detection of temporal relations among events has been mainly investigated with encoder-only models such as RoBERTa. Large Language Models (LLM) have recently shown promising performance in temporal reasoning tasks such as temporal question answering. Nevertheless, recent studies have tested the LLMs' performance in detecting temporal relations of closed-source models only, limiting the interpretability of those results. In this work, we investigate LLMs' performance and decision process in the Temporal Relation Classification task. First, we assess the performance of seven open and closed-sourced LLMs experimenting with in-context learning and lightweight fine-tuning approaches. Results show that LLMs with in-context learning significantly underperform smaller encoder-only models based on RoBERTa. Then, we delve into the possible reasons for this gap by applying explainable methods. The outcome suggests a limitation of LLMs in this task due to their autoregressive nature, which causes them to focus only on the last part of the sequence. Additionally, we evaluate the word embeddings of these two models to better understand their pre-training differences. 
The code and the fine-tuned models can be found respectively on GitHub\footnote{\url{https://github.com/BrownFortress/LLMs-TRC}}.   

\end{abstract}

\section{Introduction}
An important ability in understanding information flows such as news is to recognize the temporal relations of events, which happened, are happening, or will happen, to order them into a coherent storyline. Indeed, temporal relations are utilized in many natural language processing tasks such as narrative understanding \cite{song1988interpretation, mousavi-etal-2023-whats}, story generation \cite{han-etal-2022-generating}, summarization \cite{liu2009extractive,gung2012summarization} and temporal question answering \cite{shang2022improving, kannen-etal-2023-best}.

\begin{figure}[ht!]
\centering
\includegraphics[width=0.45\textwidth]{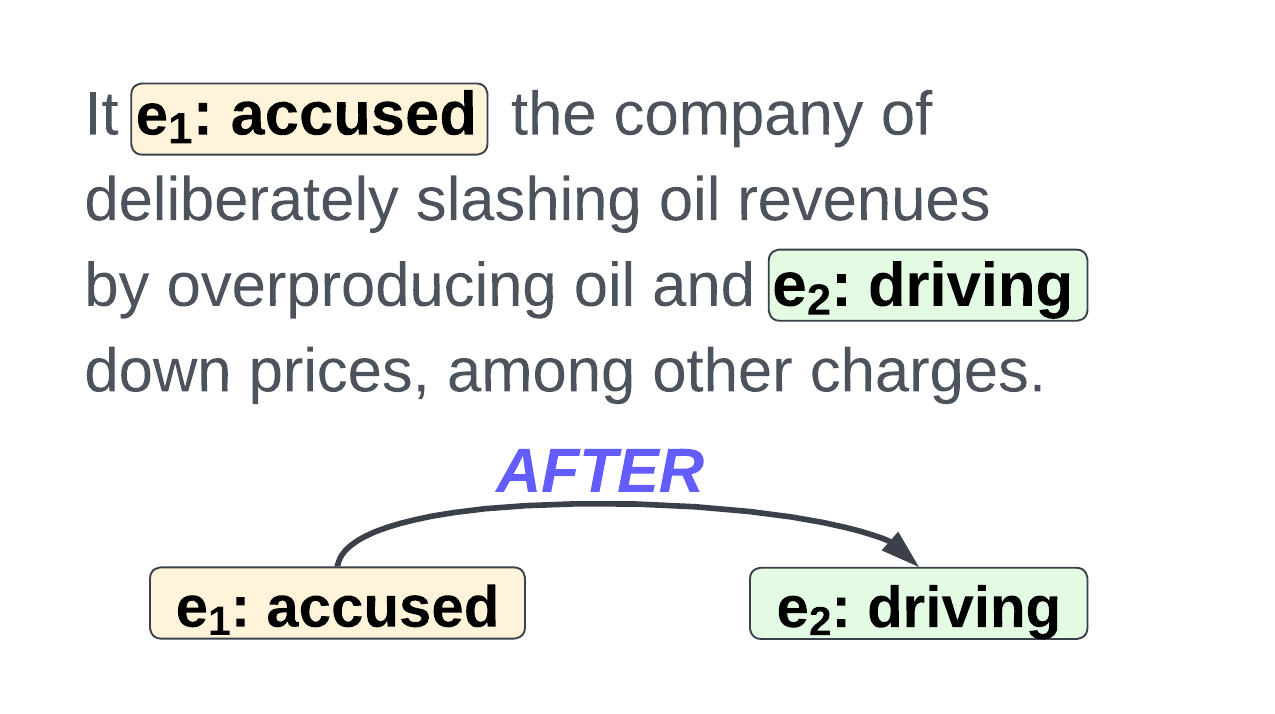}
\caption{An example taken from MATRES corpus for the Temporal Relation Classification task, in which the accusation event follows the driving event. The relation between the two event triggers, namely $e_1$:\textit{accused} and $e_2$:\textit{driving}, is annotated with a directed arc and the label \textit{AFTER}.}
\label{fig:ex1}
\end{figure}

The automatic recognition of temporal relations (e.g. \textit{before} or \textit{after}) is referred to as Temporal Relation Classification (TRC). Figure \ref{fig:ex1} depicts an example of the TRC task, as defined in the TempEval challenges \cite{verhagen-etal-2007-semeval, pustejovsky-verhagen-2009-semeval, uzzaman-etal-2013-semeval}, that is to predict the temporal relation \textit{after} between the two given connected events,  $e_1: accused$ and $e_2:driving$. The temporal relations used to annotate the corpora for training and testing models have been originally defined in Allen's interval algebra \cite{Allen1983MaintainingKA, allen1984towards}. In this, events are described as intervals rather than time points to handle explicit and implicit or vague time references. 

In recent years, several works on the TRC task have focused on exploiting a variety of features to best represent the surrounding context of the two events. Some of these are syntactic \cite{zhang2022extracting}, semantic (e.g., event arguments) \cite{zhou-etal-2022-rsgt} and discourse  \cite{mathur2021timers} features.  To utilize those features,  models have been based on complex architectures constructed on top of encoder-only pre-trained language models such as BERT \cite{kenton2019bert} and RoBERTa \cite{liu2019roberta}.  More recently, Large Language Models (LLMs) have become widely used in many natural language processing tasks, achieving state-of-the-art performance in sentiment analysis \cite{zhang2023enhancing}, named entity recognition \cite{wang2023gpt} and natural language inference \cite{chowdhery2023palm}. Although LLMs have been evaluated on the temporal question answering task \cite{wei-etal-2023-menatqa, gupta-etal-2023-temptabqa, dhingra-etal-2022-time}, in which temporal relations are implicitly used, limited studies have been conducted on the performance of LLMs on the TRC task \cite{li-etal-2023-open, yuan-etal-2023-zero, Chan2023ChatGPTE} experimenting with closed-sourced LLMs, limiting interpretability studies. 

In this paper, we study the performance and the decision process of seven open and closed-sourced Large Language Models (LLM) in performing the task of Temporal Relation Classification (TRC) over three different publicly available benchmark corpora. Along with an example-label in-context learning (ICL) approach \cite{brown2020language}, we cast the TRC task into a Question Answering task form to create QA prompts.  Furthermore, we use the Low-Rank Adaptation (LoRA) \cite{hu2021lora} technique to fine-tune Llama2 7B and  Llama2 13B to measure the upper bound performance of such models. The results show that although the autoregressive LLMs with QA prompts perform better than example-label prompts, they struggle with the TRC task compared to smaller encoder-only models based on RoBERTa in all settings. We further investigate the possible reasons for this by analyzing the most contributing tokens to the prediction extracted with KernelShap \cite{lundberg2017unified} algorithm as an XAI attribute scorer. This analysis shows that LLMs tend to focus more on the last tokens of the target sentence due to their autoregressive nature, rather than using the entire context as encoder-only models do. We evaluate the word embeddings extracted from LLMs and RoBERTa to highlight the pre-training differences between these two models. We observe that RoBERTa still performs better than LLMs, suggesting that the gap between these two models is probably due to their different pre-training strategies.

The contributions of the paper are the following:
\begin{itemize}
    \item Evaluation of seven LLMs including open and closed-sourced models with different parameter sizes and with ICL and LoRa approaches;
   \item Explainability studies on LLMs and RoBERTa models to understand the differences between the two models in their decision processes;
    \item Word embeddings evaluation and comparison between LLMs and RoBERTa model;
\end{itemize}

The remainder of the paper is organized as follows. Section \ref{sec:related_works} reviews the related works. In section \ref{sec:task_definition}, we formally define the TRC task.  Section \ref{sec:model_desc} presents the encoder-only model based on RoBERTa and the prompts for LLMs. In section \ref{sec:evaluation}, we present the results of the tested models and the explainability studies. In section \ref{sec:error_analysis}, we present and discuss the error analysis. Finally, we present our conclusions.

\section{Related works}
\label{sec:related_works}
\textbf{Temporal Relation Classification task}
Temporal Relation Classification (or Extraction) models have predominantly used encoder-based pre-trained language models, such as BERT \cite{devlin-etal-2019-bert} and RoBERTa \cite{liu2019roberta}, as a backbone. In particular, some studies have employed graph neural networks initialised with BERT and RoBERTa embeddings to model the semantic and syntactic context surrounding the events \cite{mathur2021timers, zhou-etal-2022-rsgt, zhang2022extracting}.  More recently, \citet{cohen-bar-2023-temporal}  have been fine-tuned a RoBERTa model for answering interval relation reasoning questions to predict a given temporal relation class.  Instead, LLMs have been tasked to answer multiple choice questions \cite{Chan2023ChatGPTE}, which challenges them to understand the semantics of the different temporal relations classes.

\textbf{Temporal QA}
The purpose of extracting temporal relations among events proposed in TimeML \cite{pustejovsky2003timeml} is to improve the performance of temporal Question Answering task \cite{llorens-etal-2015-semeval, meng2017temporal, yang-etal-2023-upon}, which is to answer temporal grounded queries such as \textit{``Which is the current US president?''}. Recently, the temporal QA task has been used for testing and challenging the temporal reasoning ability of LLMs \cite{wei-etal-2023-menatqa, tan-etal-2023-towards} by querying the LLMs parametric knowledge with time-grounded questions.

\textbf{Temporal Relations Corpora}
One of the first and largest corpora annotated with temporal relations is the TimeBank \cite{pustejovsky2003timebank} corpus. This corpus has been annotated using the Time-ML scheme \cite{pustejovsky2003timeml}. This scheme provides a definition used to identify events and defines a set of thirteen temporal relations which follow in principle the thirteen interval relations of Allen's interval algebra \cite{Allen1983MaintainingKA, allen1984towards}. Adaptations and refinements of the ISO Time-ML \cite{pustejovsky-etal-2010-iso} scheme,  mainly regarding the event definition and the number of relations,  have been used to annotate most of the available corpora such as in Thyme-TimeML \cite{styler-iv-etal-2014-temporal}, TimeBank-Dense \cite{cassidy2014annotation}, RED \cite{ogorman-etal-2016-richer}, MATRES, MAVEN-ERE \cite{wang-etal-2022-maven} and TIMELINE.

\section{Task Definition}
\label{sec:task_definition}
The Temporal Relation Classification (TRC) task can be defined as follows. The corpus comprises a set of documents $\mathcal{D}$. A document  $d \in \mathcal{D}$ is defined as a sequence of sentences $d=[s_1,...,s_n]$, where a sentence is a sequence of tokens i.e. $s = [w_1, ...,w_n]$. A sentence is delimited by a full stop, exclamation or question mark.  Each document in the corpus contains a set of annotated event triggers $E =\{e_1, ..., e_n\}$ where $e$ is a span of tokens of a sentence of a document i.e. $e = [w_i,...,w_j] \in s \; with  \; i>0, \; j \leq |s| \; and \; s \in d$. The TRC task is to assign a temporal relation $r$ from a predefined set $\mathcal{R}$ to a given pair of connected events $(e_i, e_j)$, where $e_i \neq e_j$. The set of relations $\mathcal{R}$ changes depending on the annotation scheme of the corpus as described in Section \ref{sec:dataset}. Besides, the temporal relations $r \in (e_i, e_j)$ and $r^{'} \in (e_j, e_i)$ are always the opposite (e.g. \textit{before} and \textit{after}) i.e. $r = \neg r^{'}$ except when $r$ is the relation \textit{equal}, where by definition $r = r^{'}$. Indeed, in Allen's interval algebra, the 13 temporal relations are composed of \textit{equal} plus six temporal relations and their corresponding opposites.

\section{Models}
\label{sec:model_desc}
In this section, we describe the encoder-only model and LLMs with fine-tuning and prompting approaches tasked with Temporal Relation Classification (TRC).

\subsection{Encoder-only Architecture} 
In this work, we use RoBERTa \cite{liu2019roberta} as an encoder-only model. This model has been pre-trained on the Masked Language Modelling (MLM) \cite{devlin-etal-2019-bert} task.  In this, the model is tasked to predict a masked token attending to the rest of the sequence.  

Inspired by \cite{zhou-etal-2022-rsgt}, we have used the following architecture to put RoBERTa in place.
The input of the models is the corresponding sentences containing the event pairs $(e_i, e_j)$. The events can be in the same (intra-sentence) or in two different (inter-sentence) sentences. Formally, the input for intra-sentence events is $C = s_k, \; s_k \in d$ where $ e_i, e_j \in s_k$ and for inter-sentence events is $C = s_i \oplus s_j, \; s_i, s_j \in d$ where $e_i \in s_i, e_j \in s_j$ .  For the latter, we concatenated the two sentences with a white space. The input $C$ is fed into a pre-trained model to compute the word embeddings of input tokens. From these, we retrieve the embedding corresponding to the tokens of the two events. Then, the event embedding is created by aggregating all the relative sub-tokens generated by the byte pair encoding tokenizer \cite{sennrich2016neural} using the max pooling function. Aggregating all tokens is important since the verb tense is a relevant aspect of this task; therefore, the \textit{-ed} sub-token (i.e. the last sub-token) can be an important feature for the event embedding. The two events embedding are then concatenated. Finally, the resulting concatenated vector is fed into a feed-forward linear layer followed by a softmax to make the prediction.
 
\subsection{In-Context Learning and Fine-Tuning}
\begin{table}[ht!]
\centering
\begin{tabular}{l|p{5cm}}
    \toprule
     \textbf{Temp. Rel.} & \textbf{Questions} \\
     \toprule
      Before & Does $e_i$ happen \textbf{\textit{before}} $e_j$?\\
      \hline
      After & Does $e_i$ happen \textbf{\textit{after}} $e_j$?\\
      \hline
      Equal & Does $e_i$ happen \textbf{\textit{at the same time as}} $e_j$?\\
      \hline
      Includes & Does $e_i$ \textbf{\textit{temporally include}}  $e_j$?\\
      \hline
      Is Included & Is $e_i$ \textbf{\textit{temporally included in}}  $e_j$?\\  
    \bottomrule
    \end{tabular}
    \caption{Casting of TRC task to QA task. The two events are identified as $e_i$ and $e_j$. In the actual prompt used for LLMs, we surrounded the two events with the tags \texttt{[event1]}\texttt{[/event1]} for $e_i$ and \texttt{[event2]}\texttt{[/event2]} for $e_j$.}
\label{tab:templates}
\end{table}

Large Language Models (LLM) have been pre-trained on a large scale of data using the autoregressive language model \cite{roth2000learning,brown2020language} as the objective task.  Differently from MLM, the model has to predict the next token $t_{k+1}$ using the previous tokens, i.e., the context sequence $t_0, ..., t_{k}$ only.

To evaluate LLMs, we experiment with in-context learning (ICL) and fine-tuning approaches \cite{brown2020language}. In the ICL experiments, we evaluate the ability of the model to understand and, thus, tackle the task by using the pre-trained knowledge only. Moreover, by updating this knowledge with fine-tuning, we measure the upper-bound performance of such models.  

Inspired by \cite{brown2020language}, we have translated the TRC task into a text-to-text task using the widely used example-label pattern, henceforth referred to as \textit{\textbf{P}}. In particular, the example in \textit{\textbf{P}} is composed of the context $C$, i.e. the concatenation of the sentences containing the events as for the RoBERTa-based model, with the two events highlighted using two tags ( ``\texttt{[event1]}$e_i$\texttt{[/event1]}'' and ``\texttt{[event2]}$e_j$\texttt{[/event2]}''). This is followed by the symbol ``->''  and the target label. Thus, given the context, the model has to generate one of the temporal relation labels, i.e. $r \in \mathcal{R}$.

We have translated the TRC task into a question answering task to  further investigate LLMs' reasoning capabilities. Motivated by LLMs' training on massive amounts of web-based data, we have designed questions that can be answered without prior knowledge about the temporal relation theory and/or formal annotation guidelines. Indeed, previous works have designed questions by involving interval reasoning \cite{li-etal-2023-open,cohen-bar-2023-temporal}.  For instance, in those to identify the \textit{before} relation the two following questions are asked "\textit{Does event $e_1$ start before $e_2$?}" and "\textit{Does event $e_1$ end before $e_2$?}".  In this work, we have formulated one question for each temporal relation class i.e. \textit{before, after, equal, includes} and \textit{is included}. These questions are listed in Table \ref{tab:templates}.  To let the model answer the question, we use the same context $C$  of the prompt \textit{\textbf{P}}.   Moreover, we have experimented with asking the model one question at a time \textit{\textbf{QA\textsubscript{1}}} and all the questions in a sequence \textit{\textbf{QA\textsubscript{2}}}. In \textit{\textbf{QA\textsubscript{2}}}, the model can use the generated response as additional context to answer the remaining questions.

We used the Low-Rank Adaptation (LoRA) technique \cite{hu2021lora} to fine-tune the LLMs, achieving an upper bound in the performance. LoRA is an efficient fine-tuning approach because it adds and trains only a small set of parameters (i.e. less than 1\% of all parameters) to the model.

\section{Experimental settings}

\subsection{Datasets}
\label{sec:dataset}
We have tested our models on TimeBank(TB)-Dense \cite{cassidy2014annotation} and MATRES \cite{ning2018multi}, which are widely used benchmarks, and TIMELINE \cite{alsayyahi2023timeline}, which is a newly released corpus. TB-Dense is composed of 36 news articles, a subset of the TimeBank corpus, published in 1990 and 1998. TB-Dense has been annotated with 6 temporal relations i.e. \textit{before}, \textit{after}, \textit{includes}, \textit{is included}, \textit{simultaneous}, and \textit{vague}. MATRES includes all the 275 news articles used in the TempEval-3 challenge. All the news articles in the train and validation sets were written and published in the time range between 1990 and 2000, while in the test set, they are all dated 2013. The corpus has been annotated with four temporal relations \textit{before}, \textit{after}, \textit{equal} and \textit{vague}.  TIMELINE is composed of 48 news articles published between 2020 and 2021 and has adopted the same temporal relation scheme of MATRES. We have used official train, development and test sizes and the label distributions are shown in Appendix \ref{sec:appendix}.

\subsection{Evaluation metrics}
All the models are evaluated using the micro-f1 score. Following the decision made for TIMELINE \cite{alsayyahi2023timeline}, we have completely removed from MATRES and TB-Dense the \textit{vague} class. This is because we want to focus only on temporal relations. The class \textit{vague} is not a temporal relation \cite{wen-ji-2021-utilizing,zhou-etal-2022-rsgt} as it has been used to handle ambiguities and disagreement during the annotation process (a.k.a. catch-all class). Indeed, we have used the class \textit{vague} to map the examples for which the LLMs output gibberish or produce contradictory responses.

In TB-Dense in the prompts \textit{\textbf{QA\textsubscript{1}}}  and \textit{\textbf{QA\textsubscript{2}}} for the label \textit{simultaneous}, we have used the same question for the temporal relation \textit{equal} as they have the same meaning.

\begin{table*}[!ht]
\centering
\renewcommand{\arraystretch}{1.2}
\begin{tabular}{p{3.2cm} |ccc|ccc|ccc}
    \toprule
     \multicolumn{1}{c|}{\textbf{Models}} & \multicolumn{3}{c|}{\textbf{MATRES}}  & \multicolumn{3}{c|}{\textbf{TIMELINE}} &  \multicolumn{3}{|c}{\textbf{TB-Dense}}\\
     \toprule
     & \multicolumn{1}{c}{\textit{\textbf{P}}}&\textit{\textbf{QA\textsubscript{1}}}&\textit{\textbf{QA\textsubscript{2}}}&{\textit{\textbf{P}}}&\textit{\textbf{QA\textsubscript{1}}}&\textit{\textbf{QA\textsubscript{2}}}&{\textit{\textbf{P}}}&\textit{\textbf{QA\textsubscript{1}}}&\textit{\textbf{QA\textsubscript{2}}}\\
     \toprule
      
      Mistral 7B &30.0&14.8&52.9& 28.7&8.1 &39.9& 5.0 & 0.4 & 0.0\\
      Mixtral 8$\times$7B &27.7&28.1&58.0& 36.1&30.2& 53.2 & 8.5 & 12.3 & 13.1\\
      Llama2 7B &31.2&14.8&56.3&41.8&9.7&58.1 & 21.7 & 1.6 & 0.6\\
      Llama2 13B & 36.7 & 8.5 & 31.1 & 41.8 & 8.0 & 28.3 & 27.9 & 3.3 & 24.3 \\
      Llama2 70B &36.6&37.0&\textbf{65.3}&39.4&48.0&\textbf{62.5} & 27.1 & 9.3 & \textbf{31.4}\\

       GPT-3 &54.0&8.0 &55.6&7.0&20.3&57.3 & 2.7 & 2.5 & 0.5 \\
       GPT-3.5 &41.2&29.6&61.2&11.7&12.2&58.5 & 19.0 & 24.6 & 12\\
        \bottomrule
       Llama2 7B\textsubscript{Fine-tuned} & 71.4 & 77.2 & 82.0 & 57.2 & \textbf{76.9} & 55.9 & 45.0 & 4.7 & 49.3 \\
       Llama2 13B\textsubscript{Fine-tuned} & 76.5 & 81.6 & \textbf{84.3} & 61.3 & 30.5 & 41.5 & \textbf{55.4} & 3.7 & 48.7 \\
        \toprule
        \toprule
        \multicolumn{1}{c|}{RoBERTa} & \multicolumn{3}{c|}{87.6}  & \multicolumn{3}{c|}{87.9} &  \multicolumn{3}{|c}{83.1}\\

    \end{tabular}
    \caption{Results achieved by LLMs with in-context learning (ICL) and fine-tuning on MATRES \cite{ning2018multi}, TB-Dense \cite{cassidy2014annotation} and TIMELINE \cite{alsayyahi2023timeline} corpora. \textit{\textbf{P}} refers to the example-label prompt. In \textit{\textbf{QA\textsubscript{1}}} and \textit{\textbf{QA\textsubscript{2}}} the TRC task is cast into two QA prompts. In \textit{\textbf{QA\textsubscript{1}}} the model answers one question at a time, while in \textit{\textbf{QA\textsubscript{2}}} the model uses as context its responses by answering the question in sequence. The results in bold are the best achieved among LLMs with ICL and fine-tuning.}
\label{tab:llms_performance}
\end{table*}

\section{Evaluation and Results}
\label{sec:evaluation} 
We have experimented with seven Large Language Models (LLM): five open-source, namely Llama2 7B, Llama2 13B, Llama2 70B \cite{touvron2023llama}, Mistral 7B \cite{jiang2023mistral}, Mixtral 8$\times$7B \cite{jiang2024mixtral}, and two closed-sourced, namely GPT-3 \cite{brown2020language} and GPT-3.5\footnote{GPT-3 is davinci-002 and GPT-3.5 is gpt-3.5-turbo-0125} \cite{chatgpt2023}. As described in section \ref{sec:model_desc}, we have adopted in-context learning (ICL) and fine-tuning approaches using the following prompts:
\begin{itemize}
    \item \textit{\textbf{P}}: given two events and the corresponding context, i.e. the sentences including the events, the model generates the label (e.g., \textit{before} or \textit{after}) that identifies a temporal relation. 
    \item \textit{\textbf{QA\textsubscript{1}}}: given two events, the corresponding context, and a question for each class (shown in Table \ref{tab:templates}),  the model answers one question at a time with \textit{yes} or \textit{no}.
    \item \textit{\textbf{QA\textsubscript{2}}}: given the same context as in  \textit{QA\textsubscript{1}},  the model answers all questions in sequence. In this setting, the generated responses become part of the context used to answer the following question.
\end{itemize}

Regarding the ICL experiments, we have experimented with zero and different numbers of few-shot examples on Llama2 7B. We have observed that the models achieve the best performance by using one example for each class of the corpus, i.e. five for TB-Dense and three for both MATRES and TIMELINE.  Furthermore, to measure the impact on the performance of the few-shot example selection, we have sampled and frozen five sets of few-shot examples to create the context for all three prompt types. 

The results \footnote{ICL results have been averaged over the five prompts; fine-tuning results have been averaged over 
five runs.} of ICL and fine-tuning experiments are reported in Table \ref{tab:llms_performance}. The prompt \textit{\textbf{P}} is effective for GPT-3 and GPT-3.5 on MATRES only. Moreover,  while on  MATRES and TIMELINE  overall, the models achieve the worst results with \textit{\textbf{QA\textsubscript{1}}}, the best results are achieved using the prompt \textit{\textbf{QA\textsubscript{2}}}. One reason for this is that in the 12\% of event-pair predictions, on average, the models with \textit{\textbf{QA\textsubscript{1}}} generate contradictory responses, i.e. answering \textit{yes} to more than one question. Adding the generated answers to the context for the next question, i.e. the prompt \textit{\textbf{QA\textsubscript{2}}}, zeroes the contradictory responses. Furthermore, Llama2 70B outperforms all the other models with  \textit{\textbf{QA\textsubscript{2}}} prompt on all corpora. Besides, all the LLMs struggle with TB-Dense, probably due to a higher number of classes to predict. 

Regarding the performance of individual LLMs with ICL, Llama2 70B outperforms all other open and closed-source models on all corpora. Furthermore, despite the 175B (billions) parameters, GPT-3 yields worse results compared with Llama2 7B on all corpora and Mixtral 8$\times$7B on MATRES whose numbers of parameters are 7B and 12B\footnote{Mixtral 8$\times$7B has 58 billion of parameters but at inference time it automatically selects and utilizes a subset of 12 billion.} respectively.  Furthermore, Llama2 7B outperforms Mistral 7B on all corpora and Mixtral 8$\times$7B on TIMELINE and TB-Dense.  Besides, LLama2 13B underperforms LLama2 7B in all corpora but TB-Dense. 

\begin{figure}[t]
    \centering
    \includegraphics[scale=0.5]{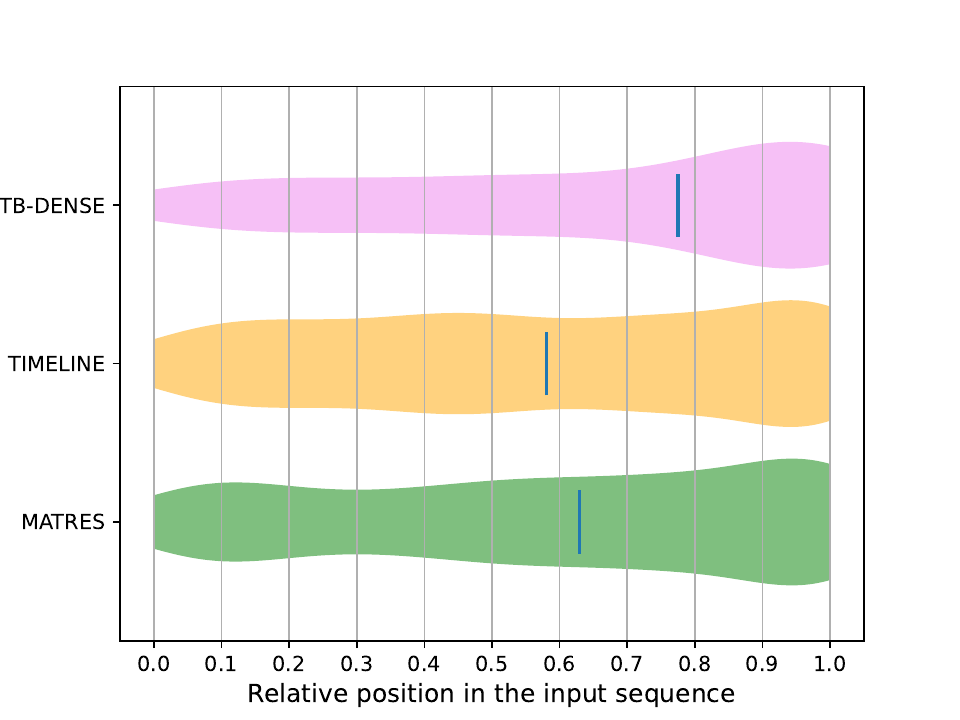}
    \caption{Distribution of the five tokens for each input sequence with the highest attribute score computed with \textit{Llama2 7B} \cite{touvron2023llama} based on the input sequence. Corpora on the y-axis and relative position in the input sequence on the x-axis. The blue line is the median.}
    \label{fig:llm_prompt_violin}
\end{figure}

To estimate the upper bound performance of LLMs, we have fine-tuned  Llama2 7B  and 13B on the three corpora using the same prompts of ICL but with a zero-shot approach. On MATRES, Llama2 13B fine-tuned using prompt  \textit{\textbf{QA\textsubscript{2}}}  achieves close results to the encoder-only model based on RoBERTa . Instead, Llama2 7B fine-tuned with \textit{\textbf{QA\textsubscript{1}}} scores the highest micro-F1 among LLMs on TIMELINE but is 11.0\% inferior to the RoBERTa-based model. On TB-Dense, while achieving the best score compared with prompt \textit{\textbf{P}}, Llama2 13B scores the highest gap of 27.7\%  w.r.t. RoBERTa-based model. A possible reason for this is that TB-Dense has two additional temporal relations compared to MATRES and TIMELINE, making the training and inference more challenging.  

Overall, the results achieved by LLMs in all settings are always outperformed by the encoder-only model based on RoBERTa. Indeed, RoBERTa scores improvements w.r.t ICL best models of 22.3\% on MATRES, 25.4\% on TIMELINE and 51.7\% on TB-Dense. Although fine-tuning Llama2 7B and 13B substantially reduces this gap on MATRES, the differences are still high in the other two datasets. 

\begin{figure}[t]
    \centering
    \includegraphics[scale=0.5]{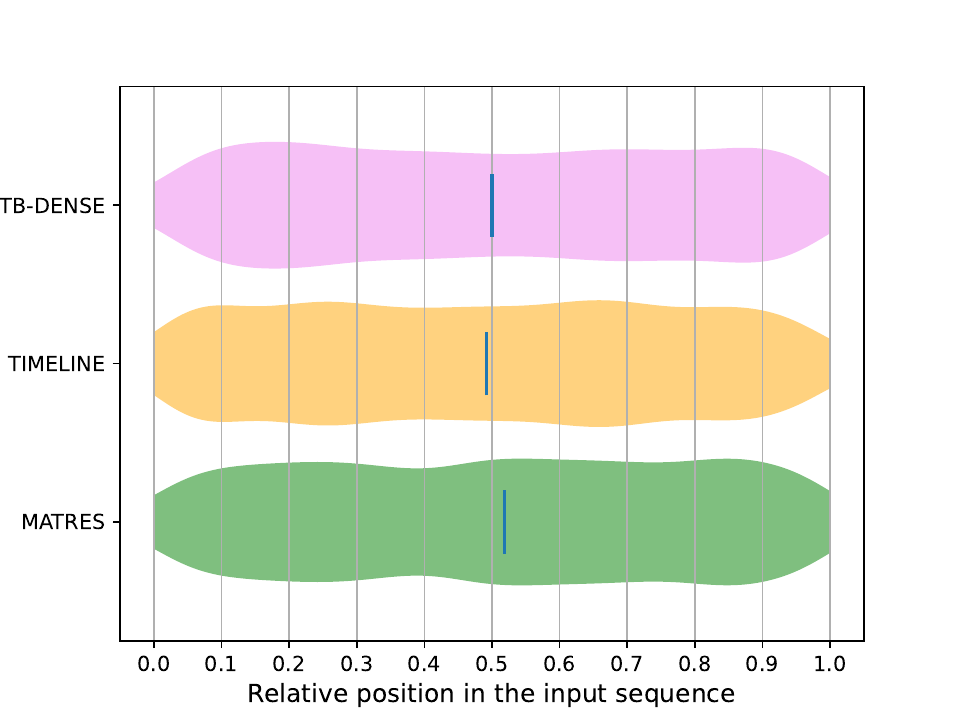}
     \caption{Distribution of the five tokens for each input sequence with the highest attribute score computed with \textit{RoBERTa} \cite{liu2019roberta} based on the input sequence. Corpora on the y-axis and relative position in the input sequence on the x-axis. The blue line is the median.}
    \label{fig:bert_violin}
\end{figure}

\begin{table*}[!ht]
\centering
\renewcommand{\arraystretch}{1.2}
\begin{tabular}{p{3.2cm} |ccc||ccc}
    \toprule
     \multicolumn{1}{c|}{\textbf{Models}} & \multicolumn{3}{c||}{\textbf{Frozen Encoder}}  & \multicolumn{3}{c}{\textbf{Full Fine-Tuning}} \\
     \toprule
     & MATR. & TIMEL. & TB-D. & MATR. & TIMEL. & TB-D.\\
     \toprule
      
      Llama2 7B &75.2&64.8&68.0&79.4&64.9&77.3 \\
      Llama2 13B &76.6&66.6&68.9&82.8&69.8&77.7 \\
      Llama2 70B &75.9&\textbf{69.1}&65.7&81.5&67.2&72.4 \\
      RoBERTa&\textbf{80.5}&65.7&\textbf{71.4}&\textbf{87.6}&\textbf{87.9}&\textbf{83.1}\\
    \bottomrule
     
    \end{tabular}
    \caption{Performance comparison between RoBERTa-based model (last row), and Llama2 7B, 13B,  and 70B. \textit{Frozen Encoder} reports the micro-F1 scores in percentage achieved by training the classification layer only. \textit{Full Fine-Tuning} reports the results attained by fine-tuning also the encoder model. \texttt{(MATR.=MATRES, TIMEL.=TIMELINE,TB-D.=TB-Dense)}}  
\label{tab:comparison}
\end{table*}
\subsection{Explainability studies}
We have studied the gap in performance between LLMs and the encoder-only RoBERTa-based model using an attribution method called KernelShap \cite{lundberg2017unified}.  KernelShap is an additive feature attribution method based on Linear LIME \cite{mishra2017local} and  SHAP values \cite{lundberg2017unified}, which gives a score to each input vector element based on its importance to the prediction. 

From each input sequence in three test sets, we have computed \footnote{For this, we have used \href{https://captum.ai}{Captum} library.} and extracted the five tokens with the highest attribution score. Regarding Llama2 7B with prompt \textit{\textbf{P}}, we have observed that 70\% of these tokens are positioned in the few-shot context, while the remaining come from the target sequence (i.e. the context of the two events to make a prediction). To be comparable with RoBERTa model, we present the distribution of the tokens based on their position coming from the target sequence only. To do this, we have computed the relative positions, i.e. scaling into a 0 to 1 range, dividing them by the sequence length. 

Figure \ref{fig:llm_prompt_violin} and Figure \ref{fig:bert_violin} show the violin plots of the distribution of the tokens with the highest attribute score based on their positions computed using Llama2 7B and RoBERTa respectively. Regarding Llama2 7B, most of the tokens with the highest attribution score are at the end of the sequence, meaning that the model tends to use only the last few tokens to make a prediction. Conversely, the distribution of the encoder-only model is more uniform, meaning that the decision process of RoBERTa considers the entire sequence. This suggests that one of the reasons behind the gap in the performance between these two kinds of models is due to the different pre-training tasks,i.e., the masked language model task \cite{devlin-etal-2019-bert} for Roberta and the autoregressive language model task for the LLMs. 

\begin{table*}[!ht]
\centering
\renewcommand{\arraystretch}{1.2}
\begin{tabular}{p{3.2cm} |ccc|ccc|ccccc}
    \toprule
     \multicolumn{1}{c|}{\textbf{Models}} & \multicolumn{3}{c|}{\textbf{MATRES}}  & \multicolumn{3}{c|}{\textbf{TIMELINE}} & \multicolumn{5}{c}{\textbf{TB-Dense}} \\
     \toprule
     & Bef & Aft & Eq & Bef & Aft & Eq & Bef & Aft & Eq & Incl & Is\_Incl \\
     \toprule
      
      Llama2 7B & 71.9 & 6.4 & 2.1 & 70.1 & 25.7 & 0.0 & 0.2 & 30.1 & 0.3 & 9.5 & 0.0 \\
      Llama2 13B & 2.0 & 53.5 & 0.7 & 5.0 & 57.7 & 2.5 & 10.6 & 41.9 & 3.3 & 11.0 & 5.1 \\
      Llama2 7B\textsubscript{Fine-tuned} & 87.2 & 77.9 & 0.0 & 81.9 & 81.2 & 0.0 & 63.7 & 34.6 & 0.0 & 0.0 & 0.0 \\
      Llama2 13B\textsubscript{Fine-tuned} & 88.6 & 82.1 & 0.0 & 68.3 & 51.8 & 0.0 & 65.8 & 51.9 & 0.0 & 0.0 & 0.0 \\
      \bottomrule
      RoBERTa & 91.8 & 86.2 & 0.0 & 89.8 & 87.4 & 8.9 & 88.6 & 86.7 & 0.0 & 56.6 & 59.2 \\
      \bottomrule
     
    \end{tabular}
    \caption{ Comparison of F1-scores in percentage for each class of each dataset achieved using the best prompt settings.
        \texttt{(Bef=Before, Aft=After, Eq=Equal,  Incl=Includes, Is\_Incl=Is Included)}
    }
\label{tab:f1_scores}
\end{table*}

\subsection{Word Embedding analysis}
We have compared the performance of the word embeddings generated by LLMs and RoBERTa models in the TRC task to investigate the differences due to the pre-training strategies. 

To do this, we have used the architecture presented in Section \ref{sec:model_desc} and replaced the encoder, i.e., RoBERTa, with Llama2 7B, 13B and 70B. We have experimented with training only the classification layer, i.e. freezing the weights of the encoder, and with full fine-tuning, using LoRa for the LLMs, to attain the upper-bound performance. 

The results of these experiments are presented in Table \ref{tab:comparison}. The micro-F1 scores attained with LLMs by training the classification layer only are higher than those achieved with the same models with ICL. However, RoBERTa still achieves the highest performance on MATRES and TB-Dense. Interestingly, LLama2 70B outperforms RoBERTa on TIMELINE. The possible reason for this is that TIMELINE contains many news articles related to the recent COVID-19 pandemic. Indeed, in more than 30\% of target sequences in the training and test sets, there is one of the following words \textit{covid-19}, \textit{coronavirus}, \textit{pandemic} and \textit{vaccine}. Considering that RoBERTa was pre-trained in 2019, those tokens are out-of-vocabulary tokens for the model. 

By further training the encoder, the models generally increase the performance on all corpora with an average improvement of 4\% and 9\%  results on MATRES and TB-Dense respectively.  Conversely, on TIMELINE, the improvement of LLama2 7B and 13B  is considerably contained, and we observe a worsening in the performance of Llama2 70B. While RoBERTa gains an increment of 22.2\%, providing additional evidence of the initial high presence of OOV tokens for the RoBERTa model.  

Overall, the results suggest that the word embeddings yielded by RoBERTa are more effective in the TRC task, supporting the outcome of the explainability studies for which one of the probable reasons for the performance gap is in the different pre-training tasks.

\section{Error Analysis}
\label{sec:error_analysis}
\begin{table}[htpb]
    \centering
    \setlength\tabcolsep{2.3pt}
    \renewcommand{\arraystretch}{1.2}
    \begin{tabular}{p{3.8cm}|c|c}
         \textbf{Models} & \textbf{Intra-sent.} & \textbf{Inter-sent.}\\
         \toprule
         RoBERTa & 85.8 & 89.6\\
         Llama2 13B\textsubscript{Fine-tuned} & 82.3 & 85.5 \\
         Llama2 70B  & 59.9 & 69.2 \\
    \end{tabular}
    \caption{Performance (micro-f1) of intra and inter-sentence event pairs  on MATRES. Intra-sentence the event pairs  are in the same sentence, while in inter-sentence they are in two different sentences. The distribution of intra and inter-sentences are 39.0\% and 61.0\% respectively.}
    \label{tab:intra_inter}
\end{table}

We have analyzed the error of the RoBERTa-based model, Llama2 70B and Llama2 13B fine-tuned by comparing the performance between intra and inter-sentences on MATRES as reported in Table \ref{tab:intra_inter}.  The encoder-only RoBERTa-based model achieves the highest intra- and inter-sentence performance. Besides, all three models underperform on the intra-sentences, where the highest difference is measured on Llama2 70B.

To investigate whether there is a subset of the test set with challenging examples for all the models, we have computed the intersection between the errors of the RoBERTa-based model and the correct predictions of Llama2 13B fine-tuned and Llama2 70B with ICL. The sizes of these intersections are 23.3\% for Llama2 13B fine-tuned and 33.7\% for Llama2 70B, which account for 2.8\% and 4.0\% of the test set, respectively. By manually inspecting these subsets, we have found that the errors are mainly due to misunderstanding of verb tenses such as past perfect continuous and future. While in the remaining errors of RoBERTa, we observed that there are challenging examples also for humans, as they require additional common sense knowledge and strong reasoning capabilities such as simulation reasoning \cite{tamari-etal-2020-language}. Some examples of such cases are shown in Table \ref{tab:challenging_examples} in Appendix \ref{sec:appendix}.

Regarding the impact of the selection of few-shot examples in ICL on LLMs, we observe that the standard deviation mainly depends on the type of prompt and the model as presented in  Table \ref{tab:std}  in Appendix \ref{sec:appendix}.  Notably, the model and the prompt with the lowest performance variability on average are Llama2 70B with 2.2\% and  \textit{\textbf{QA\textsubscript{2}}} with 2.7\%. In comparison, the few-shot selection has the highest impact on \textit{ \textbf{QA\textsubscript{1}}} with 5.8\% and Mistral 7B with 6.1\%. Thus, although requiring a relevant amount of time and resources, tuning the few-shot samples on the development set might boost the performance of some models and prompts. 

To better understand the performance of the models with ICL and fine-tuning, in Table \ref{tab:f1_scores} we report the F1-scores for each relation of the models using the best prompt settings\footnote{The F1 scores for all the models can be found in Table \ref{tab:f1_scores_all} in Appendix \ref{sec:appendix}.}. We can observe that LLMs with ICL are biased towards a specific class, i.e. the class \textit{after} for  LLama2 13B and \textit{before} for 7B. This is reduced with fine-tuning. Furthermore, the class \textit{equal} is mispredicted by all the models on all three corpora, RoBERTa included. A possible reason is that \textit{equal}  is always the least frequent class counting for 2\% to 4\% of the total number of examples.

\section{Conclusions}
In this work, we have evaluated seven open and closed-sourced LLMs on the Temporal Relation Classification task with an in-context learning and fine-tuning approach.  We have shown that the encoder-only RoBERTa-based model achieves the highest results compared to LLMs. Explainable studies suggest that one of the reasons for this gap is  due to the different pre-training tasks.  Finally,  considering the low performance and the huge amount of computational resources needed at fine-tuning and inference time, LLMs might not be the best option for the TRC task compared to a more accurate and low-resource demanding RoBERTa-based model. 

\textbf{Future work}
A possible future work is to further investigate the pre-training task differences, by pre-training two models on the autoregressive and masked language model tasks using the same parameter size and training set.  Another possible direction is to study a hybrid architecture to join the best performance of the RoBERTa-based models and the Large Language models such as Llama2 13B fine-tuned and Llama2 70B.

\section*{Limitations}
We could not experiment with the largest open-source model due to limited resources. Furthermore, the choice of using an additive feature attribution method rather than a gradient-based method is mainly based on computational time. Indeed, during our tests, we estimated that the total computational time for processing MATRES using the integrated gradients \cite{sundararajan2017axiomatic} method was three weeks compared to one day with KernelShap \cite{lundberg2017unified}. 

\bibliography{anthology, acl_latex}

\begin{thebibliography}{55}
\expandafter\ifx\csname natexlab\endcsname\relax\def\natexlab#1{#1}\fi

\bibitem[{Allen(1983)}]{Allen1983MaintainingKA}
James~F. Allen. 1983.
\newblock \href {https://api.semanticscholar.org/CorpusID:16729000} {Maintaining knowledge about temporal intervals}.
\newblock \emph{Commun. ACM}, 26:832--843.

\bibitem[{Allen(1984)}]{allen1984towards}
James~F Allen. 1984.
\newblock Towards a general theory of action and time.
\newblock \emph{Artificial intelligence}, 23(2):123--154.

\bibitem[{Alsayyahi and Batista-Navarro(2023)}]{alsayyahi2023timeline}
Sarah Alsayyahi and Riza~Theresa Batista-Navarro. 2023.
\newblock Timeline: Exhaustive annotation of temporal relations supporting the automatic ordering of events in news articles.
\newblock In \emph{Proceedings of the 2023 Conference on Empirical Methods in Natural Language Processing}, pages 16336--16348.

\bibitem[{Brown et~al.(2020)Brown, Mann, Ryder, Subbiah, Kaplan, Dhariwal, Neelakantan, Shyam, Sastry, Askell et~al.}]{brown2020language}
Tom Brown, Benjamin Mann, Nick Ryder, Melanie Subbiah, Jared~D Kaplan, Prafulla Dhariwal, Arvind Neelakantan, Pranav Shyam, Girish Sastry, Amanda Askell, et~al. 2020.
\newblock Language models are few-shot learners.
\newblock \emph{Advances in neural information processing systems}, 33:1877--1901.

\bibitem[{Cassidy et~al.(2014)Cassidy, McDowell, Chambers, and Bethard}]{cassidy2014annotation}
Taylor Cassidy, Bill McDowell, Nathanael Chambers, and Steven Bethard. 2014.
\newblock An annotation framework for dense event ordering.
\newblock In \emph{Proceedings of the 52nd Annual Meeting of the Association for Computational Linguistics (Volume 2: Short Papers)}, pages 501--506.

\bibitem[{Chan et~al.(2023)Chan, Jiayang, Wang, Jiang, Fang, Liu, and Song}]{Chan2023ChatGPTE}
Chunkit Chan, Cheng Jiayang, Weiqi Wang, Yuxin Jiang, Tianqing Fang, Xin Liu, and Yangqiu Song. 2023.
\newblock \href {https://api.semanticscholar.org/CorpusID:258418194} {Chatgpt evaluation on sentence level relations: A focus on temporal, causal, and discourse relations}.
\newblock \emph{ArXiv}, abs/2304.14827.

\bibitem[{Chowdhery et~al.(2023)Chowdhery, Narang, Devlin, Bosma, Mishra, Roberts, Barham, Chung, Sutton, Gehrmann et~al.}]{chowdhery2023palm}
Aakanksha Chowdhery, Sharan Narang, Jacob Devlin, Maarten Bosma, Gaurav Mishra, Adam Roberts, Paul Barham, Hyung~Won Chung, Charles Sutton, Sebastian Gehrmann, et~al. 2023.
\newblock Palm: Scaling language modeling with pathways.
\newblock \emph{Journal of Machine Learning Research}, 24(240):1--113.

\bibitem[{Cohen and Bar(2023)}]{cohen-bar-2023-temporal}
Omer Cohen and Kfir Bar. 2023.
\newblock \href {https://doi.org/10.18653/v1/2023.findings-acl.116} {Temporal relation classification using {B}oolean question answering}.
\newblock In \emph{Findings of the Association for Computational Linguistics: ACL 2023}, pages 1843--1852, Toronto, Canada. Association for Computational Linguistics.

\bibitem[{Devlin et~al.(2019)Devlin, Chang, Lee, and Toutanova}]{devlin-etal-2019-bert}
Jacob Devlin, Ming-Wei Chang, Kenton Lee, and Kristina Toutanova. 2019.
\newblock \href {https://doi.org/10.18653/v1/N19-1423} {{BERT}: Pre-training of deep bidirectional transformers for language understanding}.
\newblock In \emph{Proceedings of the 2019 Conference of the North {A}merican Chapter of the Association for Computational Linguistics: Human Language Technologies, Volume 1 (Long and Short Papers)}, pages 4171--4186, Minneapolis, Minnesota. Association for Computational Linguistics.

\bibitem[{Dhingra et~al.(2022)Dhingra, Cole, Eisenschlos, Gillick, Eisenstein, and Cohen}]{dhingra-etal-2022-time}
Bhuwan Dhingra, Jeremy~R. Cole, Julian~Martin Eisenschlos, Daniel Gillick, Jacob Eisenstein, and William~W. Cohen. 2022.
\newblock \href {https://doi.org/10.1162/tacl_a_00459} {Time-aware language models as temporal knowledge bases}.
\newblock \emph{Transactions of the Association for Computational Linguistics}, 10:257--273.

\bibitem[{Gung and Kalita(2012)}]{gung2012summarization}
James Gung and Jugal Kalita. 2012.
\newblock Summarization of historical articles using temporal event clustering.
\newblock In \emph{Proceedings of the 2012 Conference of the North American Chapter of the Association for Computational Linguistics: Human Language Technologies}, pages 631--635.

\bibitem[{Gupta et~al.(2023)Gupta, Kandoi, Vora, Zhang, He, Reinanda, and Srikumar}]{gupta-etal-2023-temptabqa}
Vivek Gupta, Pranshu Kandoi, Mahek Vora, Shuo Zhang, Yujie He, Ridho Reinanda, and Vivek Srikumar. 2023.
\newblock \href {https://doi.org/10.18653/v1/2023.emnlp-main.149} {{T}emp{T}ab{QA}: Temporal question answering for semi-structured tables}.
\newblock In \emph{Proceedings of the 2023 Conference on Empirical Methods in Natural Language Processing}, pages 2431--2453, Singapore. Association for Computational Linguistics.

\bibitem[{Han et~al.(2022)Han, Castro~Ferreira, and Gardent}]{han-etal-2022-generating}
Kelvin Han, Thiago Castro~Ferreira, and Claire Gardent. 2022.
\newblock \href {https://aclanthology.org/2022.lrec-1.29} {Generating questions from {W}ikidata triples}.
\newblock In \emph{Proceedings of the Thirteenth Language Resources and Evaluation Conference}, pages 277--290, Marseille, France. European Language Resources Association.

\bibitem[{Hu et~al.(2021)Hu, Shen, Wallis, Allen-Zhu, Li, Wang, Wang, and Chen}]{hu2021lora}
Edward~J Hu, Yelong Shen, Phillip Wallis, Zeyuan Allen-Zhu, Yuanzhi Li, Shean Wang, Lu~Wang, and Weizhu Chen. 2021.
\newblock Lora: Low-rank adaptation of large language models.
\newblock \emph{arXiv preprint arXiv:2106.09685}.

\bibitem[{Jiang et~al.(2023)Jiang, Sablayrolles, Mensch, Bamford, Chaplot, Casas, Bressand, Lengyel, Lample, Saulnier et~al.}]{jiang2023mistral}
Albert~Q Jiang, Alexandre Sablayrolles, Arthur Mensch, Chris Bamford, Devendra~Singh Chaplot, Diego de~las Casas, Florian Bressand, Gianna Lengyel, Guillaume Lample, Lucile Saulnier, et~al. 2023.
\newblock Mistral 7b.
\newblock \emph{arXiv preprint arXiv:2310.06825}.

\bibitem[{Jiang et~al.(2024)Jiang, Sablayrolles, Roux, Mensch, Savary, Bamford, Chaplot, Casas, Hanna, Bressand et~al.}]{jiang2024mixtral}
Albert~Q Jiang, Alexandre Sablayrolles, Antoine Roux, Arthur Mensch, Blanche Savary, Chris Bamford, Devendra~Singh Chaplot, Diego de~las Casas, Emma~Bou Hanna, Florian Bressand, et~al. 2024.
\newblock Mixtral of experts.
\newblock \emph{arXiv preprint arXiv:2401.04088}.

\bibitem[{Kannen et~al.(2023)Kannen, Sharma, Neelam, Khandelwal, Ikbal, Karanam, and Subramaniam}]{kannen-etal-2023-best}
Nithish Kannen, Udit Sharma, Sumit Neelam, Dinesh Khandelwal, Shajith Ikbal, Hima Karanam, and L~Subramaniam. 2023.
\newblock \href {https://doi.org/10.18653/v1/2023.emnlp-main.287} {Best of both worlds: Towards improving temporal knowledge base question answering via targeted fact extraction}.
\newblock In \emph{Proceedings of the 2023 Conference on Empirical Methods in Natural Language Processing}, pages 4729--4744, Singapore. Association for Computational Linguistics.

\bibitem[{Kenton and Toutanova(2019)}]{kenton2019bert}
Jacob Devlin Ming-Wei~Chang Kenton and Lee~Kristina Toutanova. 2019.
\newblock Bert: Pre-training of deep bidirectional transformers for language understanding.
\newblock In \emph{Proceedings of NAACL-HLT}, pages 4171--4186.

\bibitem[{Li et~al.(2023)Li, Zhao, Li, Ji, Callison-Burch, and Han}]{li-etal-2023-open}
Sha Li, Ruining Zhao, Manling Li, Heng Ji, Chris Callison-Burch, and Jiawei Han. 2023.
\newblock \href {https://doi.org/10.18653/v1/2023.acl-long.312} {Open-domain hierarchical event schema induction by incremental prompting and verification}.
\newblock In \emph{Proceedings of the 61st Annual Meeting of the Association for Computational Linguistics (Volume 1: Long Papers)}, pages 5677--5697, Toronto, Canada. Association for Computational Linguistics.

\bibitem[{Liu et~al.(2009)Liu, Li, and Hu}]{liu2009extractive}
Maofu Liu, Wenjie Li, and Huijun Hu. 2009.
\newblock Extractive summarization based on event term temporal relation graph and critical chain.
\newblock In \emph{Information Retrieval Technology: 5th Asia Information Retrieval Symposium, AIRS 2009, Sapporo, Japan, October 21-23, 2009. Proceedings 5}, pages 87--99. Springer.

\bibitem[{Liu et~al.(2019)Liu, Ott, Goyal, Du, Joshi, Chen, Levy, Lewis, Zettlemoyer, and Stoyanov}]{liu2019roberta}
Yinhan Liu, Myle Ott, Naman Goyal, Jingfei Du, Mandar Joshi, Danqi Chen, Omer Levy, Mike Lewis, Luke Zettlemoyer, and Veselin Stoyanov. 2019.
\newblock Roberta: A robustly optimized bert pretraining approach.
\newblock \emph{arXiv preprint arXiv:1907.11692}.

\bibitem[{Llorens et~al.(2015)Llorens, Chambers, UzZaman, Mostafazadeh, Allen, and Pustejovsky}]{llorens-etal-2015-semeval}
Hector Llorens, Nathanael Chambers, Naushad UzZaman, Nasrin Mostafazadeh, James Allen, and James Pustejovsky. 2015.
\newblock \href {https://doi.org/10.18653/v1/S15-2134} {{S}em{E}val-2015 task 5: {QA} {T}emp{E}val - evaluating temporal information understanding with question answering}.
\newblock In \emph{Proceedings of the 9th International Workshop on Semantic Evaluation ({S}em{E}val 2015)}, pages 792--800, Denver, Colorado. Association for Computational Linguistics.

\bibitem[{Loshchilov and Hutter(2017)}]{loshchilov2017decoupled}
Ilya Loshchilov and Frank Hutter. 2017.
\newblock Decoupled weight decay regularization.
\newblock \emph{arXiv preprint arXiv:1711.05101}.

\bibitem[{Lundberg and Lee(2017)}]{lundberg2017unified}
Scott~M Lundberg and Su-In Lee. 2017.
\newblock A unified approach to interpreting model predictions.
\newblock \emph{Advances in neural information processing systems}, 30.

\bibitem[{Mathur et~al.(2021)Mathur, Jain, Dernoncourt, Morariu, Tran, and Manocha}]{mathur2021timers}
Puneet Mathur, Rajiv Jain, Franck Dernoncourt, Vlad Morariu, Quan~Hung Tran, and Dinesh Manocha. 2021.
\newblock Timers: document-level temporal relation extraction.
\newblock In \emph{Proceedings of the 59th Annual Meeting of the Association for Computational Linguistics and the 11th International Joint Conference on Natural Language Processing (Volume 2: Short Papers)}, pages 524--533.

\bibitem[{Meng et~al.(2017)Meng, Rumshisky, and Romanov}]{meng2017temporal}
Yuanliang Meng, Anna Rumshisky, and Alexey Romanov. 2017.
\newblock Temporal information extraction for question answering using syntactic dependencies in an lstm-based architecture.
\newblock \emph{arXiv preprint arXiv:1703.05851}.

\bibitem[{Mishra et~al.(2017)Mishra, Sturm, and Dixon}]{mishra2017local}
Saumitra Mishra, Bob~L Sturm, and Simon Dixon. 2017.
\newblock Local interpretable model-agnostic explanations for music content analysis.
\newblock In \emph{ISMIR}, volume~53, pages 537--543.

\bibitem[{Mousavi et~al.(2023)Mousavi, Tanaka, Roccabruna, Yoshino, Nakamura, and Riccardi}]{mousavi-etal-2023-whats}
Seyed~Mahed Mousavi, Shohei Tanaka, Gabriel Roccabruna, Koichiro Yoshino, Satoshi Nakamura, and Giuseppe Riccardi. 2023.
\newblock \href {https://doi.org/10.18653/v1/2023.wnu-1.1} {What{'}s new? identifying the unfolding of new events in a narrative}.
\newblock In \emph{Proceedings of the The 5th Workshop on Narrative Understanding}, pages 1--10, Toronto, Canada. Association for Computational Linguistics.

\bibitem[{Ning et~al.(2018)Ning, Wu, and Roth}]{ning2018multi}
Qiang Ning, Hao Wu, and Dan Roth. 2018.
\newblock A multi-axis annotation scheme for event temporal relations.
\newblock In \emph{Proceedings of the 56th Annual Meeting of the Association for Computational Linguistics (Volume 1: Long Papers)}, pages 1318--1328.

\bibitem[{O{'}Gorman et~al.(2016)O{'}Gorman, Wright-Bettner, and Palmer}]{ogorman-etal-2016-richer}
Tim O{'}Gorman, Kristin Wright-Bettner, and Martha Palmer. 2016.
\newblock \href {https://doi.org/10.18653/v1/W16-5706} {Richer event description: Integrating event coreference with temporal, causal and bridging annotation}.
\newblock In \emph{Proceedings of the 2nd Workshop on Computing News Storylines ({CNS} 2016)}, pages 47--56, Austin, Texas. Association for Computational Linguistics.

\bibitem[{OpenAI(2023)}]{chatgpt2023}
OpenAI. 2023.
\newblock \href {https://chat.openai.com/chat} {Chatgpt}.

\bibitem[{Pustejovsky et~al.(2003{\natexlab{a}})Pustejovsky, Castano, Ingria, Saur{\i}, Gaizauskas, Setzer, Katz, and Radev}]{pustejovsky2003timeml}
James Pustejovsky, Jos{\'e} Castano, Robert Ingria, Roser Saur{\i}, Rob Gaizauskas, Andrea Setzer, Graham Katz, and D~Radev. 2003{\natexlab{a}}.
\newblock Timeml: A specification language for temporal and event expressions.
\newblock In \emph{Proceedings of the International Workshop of Computational Semantics}, page 193.

\bibitem[{Pustejovsky et~al.(2003{\natexlab{b}})Pustejovsky, Hanks, Sauri, See, Gaizauskas, Setzer, Radev, Sundheim, Day, Ferro et~al.}]{pustejovsky2003timebank}
James Pustejovsky, Patrick Hanks, Roser Sauri, Andrew See, Robert Gaizauskas, Andrea Setzer, Dragomir Radev, Beth Sundheim, David Day, Lisa Ferro, et~al. 2003{\natexlab{b}}.
\newblock The timebank corpus.
\newblock In \emph{Corpus linguistics}, volume 2003, page~40. Lancaster, UK.

\bibitem[{Pustejovsky et~al.(2010)Pustejovsky, Lee, Bunt, and Romary}]{pustejovsky-etal-2010-iso}
James Pustejovsky, Kiyong Lee, Harry Bunt, and Laurent Romary. 2010.
\newblock \href {http://www.lrec-conf.org/proceedings/lrec2010/pdf/55_Paper.pdf} {{ISO}-{T}ime{ML}: An international standard for semantic annotation}.
\newblock In \emph{Proceedings of the Seventh International Conference on Language Resources and Evaluation ({LREC}'10)}, Valletta, Malta. European Language Resources Association (ELRA).

\bibitem[{Pustejovsky and Verhagen(2009)}]{pustejovsky-verhagen-2009-semeval}
James Pustejovsky and Marc Verhagen. 2009.
\newblock \href {https://aclanthology.org/W09-2418} {{S}em{E}val-2010 task 13: Evaluating events, time expressions, and temporal relations ({T}emp{E}val-2)}.
\newblock In \emph{Proceedings of the Workshop on Semantic Evaluations: Recent Achievements and Future Directions ({SEW}-2009)}, pages 112--116, Boulder, Colorado. Association for Computational Linguistics.

\bibitem[{Roth(2000)}]{roth2000learning}
Dan Roth. 2000.
\newblock Learning in natural language: Theory and algorithmic approaches.
\newblock In \emph{Fourth Conference on Computational Natural Language Learning and the Second Learning Language in Logic Workshop}.

\bibitem[{Sennrich et~al.(2016)Sennrich, Haddow, and Birch}]{sennrich2016neural}
Rico Sennrich, Barry Haddow, and Alexandra Birch. 2016.
\newblock Neural machine translation of rare words with subword units.
\newblock In \emph{Proceedings of the 54th Annual Meeting of the Association for Computational Linguistics (Volume 1: Long Papers)}, pages 1715--1725.

\bibitem[{Shang et~al.(2022)Shang, Wang, Qi, and Huang}]{shang2022improving}
Chao Shang, Guangtao Wang, Peng Qi, and Jing Huang. 2022.
\newblock Improving time sensitivity for question answering over temporal knowledge graphs.
\newblock In \emph{Proceedings of the 60th Annual Meeting of the Association for Computational Linguistics (Volume 1: Long Papers)}, pages 8017--8026.

\bibitem[{Song and Cohen(1988)}]{song1988interpretation}
Fei Song and Robin Cohen. 1988.
\newblock The interpretation of temporal rdations in narrative.
\newblock \emph{AAAI-88 Proceedings, at Saint Paul}.

\bibitem[{Styler~IV et~al.(2014)Styler~IV, Bethard, Finan, Palmer, Pradhan, de~Groen, Erickson, Miller, Lin, Savova, and Pustejovsky}]{styler-iv-etal-2014-temporal}
William~F. Styler~IV, Steven Bethard, Sean Finan, Martha Palmer, Sameer Pradhan, Piet~C de~Groen, Brad Erickson, Timothy Miller, Chen Lin, Guergana Savova, and James Pustejovsky. 2014.
\newblock \href {https://doi.org/10.1162/tacl_a_00172} {Temporal annotation in the clinical domain}.
\newblock \emph{Transactions of the Association for Computational Linguistics}, 2:143--154.

\bibitem[{Sundararajan et~al.(2017)Sundararajan, Taly, and Yan}]{sundararajan2017axiomatic}
Mukund Sundararajan, Ankur Taly, and Qiqi Yan. 2017.
\newblock Axiomatic attribution for deep networks.
\newblock In \emph{International conference on machine learning}, pages 3319--3328. PMLR.

\bibitem[{Tamari et~al.(2020)Tamari, Shani, Hope, Petruck, Abend, and Shahaf}]{tamari-etal-2020-language}
Ronen Tamari, Chen Shani, Tom Hope, Miriam R~L Petruck, Omri Abend, and Dafna Shahaf. 2020.
\newblock \href {https://doi.org/10.18653/v1/2020.acl-main.559} {{L}anguage (re)modelling: {T}owards embodied language understanding}.
\newblock In \emph{Proceedings of the 58th Annual Meeting of the Association for Computational Linguistics}, pages 6268--6281, Online. Association for Computational Linguistics.

\bibitem[{Tan et~al.(2023)Tan, Ng, and Bing}]{tan-etal-2023-towards}
Qingyu Tan, Hwee~Tou Ng, and Lidong Bing. 2023.
\newblock \href {https://doi.org/10.18653/v1/2023.acl-long.828} {Towards benchmarking and improving the temporal reasoning capability of large language models}.
\newblock In \emph{Proceedings of the 61st Annual Meeting of the Association for Computational Linguistics (Volume 1: Long Papers)}, pages 14820--14835, Toronto, Canada. Association for Computational Linguistics.

\bibitem[{Touvron et~al.(2023)Touvron, Martin, Stone, Albert, Almahairi, Babaei, Bashlykov, Batra, Bhargava, Bhosale et~al.}]{touvron2023llama}
Hugo Touvron, Louis Martin, Kevin Stone, Peter Albert, Amjad Almahairi, Yasmine Babaei, Nikolay Bashlykov, Soumya Batra, Prajjwal Bhargava, Shruti Bhosale, et~al. 2023.
\newblock Llama 2: Open foundation and fine-tuned chat models.
\newblock \emph{arXiv preprint arXiv:2307.09288}.

\bibitem[{UzZaman et~al.(2013)UzZaman, Llorens, Derczynski, Allen, Verhagen, and Pustejovsky}]{uzzaman-etal-2013-semeval}
Naushad UzZaman, Hector Llorens, Leon Derczynski, James Allen, Marc Verhagen, and James Pustejovsky. 2013.
\newblock \href {https://aclanthology.org/S13-2001} {{S}em{E}val-2013 task 1: {T}emp{E}val-3: Evaluating time expressions, events, and temporal relations}.
\newblock In \emph{Second Joint Conference on Lexical and Computational Semantics (*{SEM}), Volume 2: Proceedings of the Seventh International Workshop on Semantic Evaluation ({S}em{E}val 2013)}, pages 1--9, Atlanta, Georgia, USA. Association for Computational Linguistics.

\bibitem[{Verhagen et~al.(2007)Verhagen, Gaizauskas, Schilder, Hepple, Katz, and Pustejovsky}]{verhagen-etal-2007-semeval}
Marc Verhagen, Robert Gaizauskas, Frank Schilder, Mark Hepple, Graham Katz, and James Pustejovsky. 2007.
\newblock \href {https://aclanthology.org/S07-1014} {{S}em{E}val-2007 task 15: {T}emp{E}val temporal relation identification}.
\newblock In \emph{Proceedings of the Fourth International Workshop on Semantic Evaluations ({S}em{E}val-2007)}, pages 75--80, Prague, Czech Republic. Association for Computational Linguistics.

\bibitem[{Wang et~al.(2023)Wang, Sun, Li, Ouyang, Wu, Zhang, Li, and Wang}]{wang2023gpt}
Shuhe Wang, Xiaofei Sun, Xiaoya Li, Rongbin Ouyang, Fei Wu, Tianwei Zhang, Jiwei Li, and Guoyin Wang. 2023.
\newblock Gpt-ner: Named entity recognition via large language models.
\newblock \emph{arXiv preprint arXiv:2304.10428}.

\bibitem[{Wang et~al.(2022)Wang, Chen, Ding, Peng, Wang, Lin, Han, Hou, Li, Liu, Li, and Zhou}]{wang-etal-2022-maven}
Xiaozhi Wang, Yulin Chen, Ning Ding, Hao Peng, Zimu Wang, Yankai Lin, Xu~Han, Lei Hou, Juanzi Li, Zhiyuan Liu, Peng Li, and Jie Zhou. 2022.
\newblock \href {https://doi.org/10.18653/v1/2022.emnlp-main.60} {{MAVEN}-{ERE}: A unified large-scale dataset for event coreference, temporal, causal, and subevent relation extraction}.
\newblock In \emph{Proceedings of the 2022 Conference on Empirical Methods in Natural Language Processing}, pages 926--941, Abu Dhabi, United Arab Emirates. Association for Computational Linguistics.

\bibitem[{Wei et~al.(2023)Wei, Su, Ma, Yu, Lei, Zhang, Zhao, and Liu}]{wei-etal-2023-menatqa}
Yifan Wei, Yisong Su, Huanhuan Ma, Xiaoyan Yu, Fangyu Lei, Yuanzhe Zhang, Jun Zhao, and Kang Liu. 2023.
\newblock \href {https://doi.org/10.18653/v1/2023.findings-emnlp.100} {{M}enat{QA}: A new dataset for testing the temporal comprehension and reasoning abilities of large language models}.
\newblock In \emph{Findings of the Association for Computational Linguistics: EMNLP 2023}, pages 1434--1447, Singapore. Association for Computational Linguistics.

\bibitem[{Wen and Ji(2021)}]{wen-ji-2021-utilizing}
Haoyang Wen and Heng Ji. 2021.
\newblock \href {https://doi.org/10.18653/v1/2021.emnlp-main.815} {Utilizing relative event time to enhance event-event temporal relation extraction}.
\newblock In \emph{Proceedings of the 2021 Conference on Empirical Methods in Natural Language Processing}, pages 10431--10437, Online and Punta Cana, Dominican Republic. Association for Computational Linguistics.

\bibitem[{Yang et~al.(2023)Yang, Li, Bing, and Lam}]{yang-etal-2023-upon}
Sen Yang, Xin Li, Lidong Bing, and Wai Lam. 2023.
\newblock \href {https://doi.org/10.18653/v1/2023.emnlp-main.728} {Once upon a ${\it time}$ in ${\it graph}$: Relative-time pretraining for complex temporal reasoning}.
\newblock In \emph{Proceedings of the 2023 Conference on Empirical Methods in Natural Language Processing}, pages 11879--11895, Singapore. Association for Computational Linguistics.

\bibitem[{Yuan et~al.(2023)Yuan, Xie, and Ananiadou}]{yuan-etal-2023-zero}
Chenhan Yuan, Qianqian Xie, and Sophia Ananiadou. 2023.
\newblock \href {https://doi.org/10.18653/v1/2023.bionlp-1.7} {Zero-shot temporal relation extraction with {C}hat{GPT}}.
\newblock In \emph{The 22nd Workshop on Biomedical Natural Language Processing and BioNLP Shared Tasks}, pages 92--102, Toronto, Canada. Association for Computational Linguistics.

\bibitem[{Zhang et~al.(2023)Zhang, Yang, Zhou, Ali~Babar, and Liu}]{zhang2023enhancing}
Boyu Zhang, Hongyang Yang, Tianyu Zhou, Muhammad Ali~Babar, and Xiao-Yang Liu. 2023.
\newblock Enhancing financial sentiment analysis via retrieval augmented large language models.
\newblock In \emph{Proceedings of the Fourth ACM International Conference on AI in Finance}, pages 349--356.

\bibitem[{Zhang et~al.(2022)Zhang, Ning, and Huang}]{zhang2022extracting}
Shuaicheng Zhang, Qiang Ning, and Lifu Huang. 2022.
\newblock Extracting temporal event relation with syntax-guided graph transformer.
\newblock In \emph{Findings of the Association for Computational Linguistics: NAACL 2022}, pages 379--390.

\bibitem[{Zhou et~al.(2022)Zhou, Dong, Tu, Wang, and Dou}]{zhou-etal-2022-rsgt}
Jie Zhou, Shenpo Dong, Hongkui Tu, Xiaodong Wang, and Yong Dou. 2022.
\newblock \href {https://aclanthology.org/2022.coling-1.174} {{RSGT}: Relational structure guided temporal relation extraction}.
\newblock In \emph{Proceedings of the 29th International Conference on Computational Linguistics}, pages 2001--2010, Gyeongju, Republic of Korea. International Committee on Computational Linguistics.

\end{thebibliography}

\appendix

\section{Appendix}
\label{sec:appendix}
We report the tables regarding the label distribution of the three corpora (Table \ref{tab:lab_distribution}), the number of relations (Table \ref{tab:num_relations}), and the standard deviation for each in-context learning experiment done using five different few-shot prompts (Table \ref{tab:std}). Furthermore, we provide the schema of the three types of prompts that we used in our experiments in Table \ref{tab:prompts}.  Table \ref{tab:f1_scores_all}
presents the F1-scores for each class achieved using the best prompt settings. In addition to this, in Table \ref{tab:challenging_examples} we provide a couple of wrongly predicted examples by the RoBERTa-based model which is challenging also for humans.  

In this work, we have respected the original intended uses of datasets, models and any other artefacts.

\subsection{Hyperparameters}
For MATRES and TB-Dense we have chunked the text into sentences using NLTK\footnote{https://www.nltk.org} library as the corpus is natively split into paragraphs. The reason for this is to have minimal context for a given event.

Regarding the RoBERTa-based model, we experimented with different configurations and hyperparameters. In this, we have used one AdamW \cite{loshchilov2017decoupled} optimizer for the encoder and another for the feed-forward layers (i.e. the classifier) with a learning rate of 1e-5 and 1e-4 respectively. Furthermore, we have used a linear scheduler on the optimizer of the encoder with warmup steps set to 10\% of the total steps in training. In all the experiments the batch size is set to 8 event-pair data points.

In the in-context learning experiments \cite{brown2020language}, the number of few-shots, i.e. the number of ground truth examples used as context for the prediction, is set to one for each class of the corpus, i.e. five for TB-Dense and three examples for both MATRES and TIMELINE. The templates of the few-shot are replications of the prompt in the zero-shot version. An example of each different prompt can be found in Table \ref{tab:prompts}. The examples have been extracted randomly from the training set of each corpus. To measure the impact on the performance of this selection, we have sampled and frozen five different sets to create the few-shots context for all three prompt types, i.e. \textbf{\textit{P}}, \textit{\textbf{QA\textsubscript{1}}} and \textit{\textbf{QA\textsubscript{2}}}. In the few-shot context of \textit{\textbf{QA\textsubscript{1}}} and \textit{\textbf{QA\textsubscript{2}}} and at inference time \textit{\textbf{QA\textsubscript{2}}}, we have kept the same order of the questions for all models and datasets which is  \textit{after}, \textit{before}, \textit{equal} and additionally for TB-Dense \textit{includes} and \textit{is included}.  To fine-tune Llama2 7B, we have used a zero-shot approach as the model has to learn the task from the back-propagation of the error rather than the few-shots in the context. In this, we have used a learning rate of 1e-4 with AdamW \cite{loshchilov2017decoupled}  optimizer, training batch size 8, and we set the rank and alpha of LoRA to 32 and 64 respectively. 

To fine-tune and test the RoBERTa model we used one NVIDIA GPU 3090Ti with 24GB. Regarding LLMs, we used four NVIDIA GPUs A100 with 80GB. The amount of GPU time needed to run all the experiments is around one month. To test the closed-sourced LLMs we have spent around \$350 in API calls.

\begin{table}[ht!]
\centering
\setlength\tabcolsep{2.3pt}
\begin{tabular}{l|c|c|c}
    \toprule
     \textbf{Temp. Rel.} & \textbf{MATRES} & \textbf{TIMELINE} & \textbf{TB-Dense} \\
     \toprule
      Before & 58.0\%&51.0\%&42.0\%\\
      \hline
      After & 38.0\%&47.0\%&  35.0\%\\
      \hline
      Equal &4.0\% &2.0\%& 3.0\%\\
      \hline
      Includes & -& - & 9.0\%\\
      \hline
      Is Included & - & - & 11.0\%\\  
    \end{tabular}
    \caption{Label distribution of the three corpora computed on the entire three partitions (i.e. train, dev and test sets). \textit{Simultaneous} in TB-Dense is mapped to \textit{equal}.}
\label{tab:lab_distribution}
\end{table}

\begin{table}[ht!]
\centering

\begin{tabular}{l|c|c|c}
    \toprule
     \textbf{Corpus} & \textbf{Train} & \textbf{Dev} & \textbf{Test} \\
     \toprule
      MATRES & 9074 &2133&724\\
      \hline
      TIMELINE &2384&284&  685\\
      \hline
      TB-Dense &2008&375& 789\\
    
    \end{tabular}
    \caption{Number of relations over the three partitions for each corpus, without the class \textit{vague}.}
\label{tab:num_relations}
\end{table}

\begin{table*}[!ht]
\centering
\renewcommand{\arraystretch}{1.0}

\begin{tabular}{p{3.2cm} |ccc|ccc|ccc}
    \toprule
     \multicolumn{1}{c|}{\textbf{Models}} & \multicolumn{3}{c|}{\textbf{MATRES}}  & \multicolumn{3}{c|}{\textbf{TIMELINE}} &  \multicolumn{3}{|c}{\textbf{TB-Dense}}\\
     \toprule
     & \multicolumn{1}{c}{\textit{\textbf{P}}}&\textit{\textbf{QA\textsubscript{1}}}&\textit{\textbf{QA\textsubscript{2}}}&{\textit{\textbf{P}}}&\textit{\textbf{QA\textsubscript{1}}}&\textit{\textbf{QA\textsubscript{2}}}&{\textit{\textbf{P}}}&\textit{\textbf{QA\textsubscript{1}}}&\textit{\textbf{QA\textsubscript{2}}}\\
     \toprule
      
      Mistral 7B &13.9&7.4&4.7&15.3&4.0&8.5&1.0&0.1&0.3\\
      Mixtral 8$\times$7B &9.0&7.1&0.8&7.9&8.5& 4.0&11.5&1.1&1.5\\
      Llama2 7B &0.0&11.4&3.8&0.0&8.8&3.4&2.5&2.3&0.2\\
      Llama2 13B & 0.6 & 4.0 & 12.6 & 1.1 & 5.0 & 12.0 & 8.3 & 2.1 & 9.3 \\
      Llama2 70B &0.4&6.2&1.7&1.9&2.2&0.9&1.9&1.0&1.2\\
      \bottomrule
      GPT-3 &4.7&5.8&9.6&6.4&1.3&1.3&10.7&1.5&1.8\\
      GPT-3.5 &1.0&6.6&2.0&6.1&11.5&3.6&6.3&1.3&4.1\\
       
    \end{tabular}
    \caption{Standard deviation computed on ten randomly generated prompts for the results achieved by LLMs with in-context learning (ICL) on MATRES \cite{ning2018multi}, TB-Dense \cite{cassidy2014annotation} and TIMELINE \cite{alsayyahi2023timeline} corpora. \textit{\textbf{P}} refers to the example-label prompt. In \textit{\textbf{QA\textsubscript{1}}} and \textit{\textbf{QA\textsubscript{2}}} the TRC task is cast into two QA prompts. In \textit{\textbf{QA\textsubscript{1}}} the model answers one question at a time, while in \textit{\textbf{QA\textsubscript{2}}} the model uses as context its responses by answering the question in sequence.}
\label{tab:std}
\end{table*}

\begin{table*}[htpb]
    \centering
    \renewcommand{\arraystretch}{1.6}
    \begin{tabular}{c|p{12cm}}
         \textbf{Prompt types} & \textbf{Prompt}\\
         \toprule
         \textbf{\textit{P}} & \textit{Given the context:} It \texttt{[event1]} \textbf{accused} \texttt{[/event1]} the company of deliberately slashing oil revenues by overproducing oil and \texttt{[event2]} \textbf{driving} \texttt{[/event2]} down prices, among other charges. -> AFTER\\
         \hline
        \textbf{\textit{ QA\textsubscript{1}}} & \textit{Given the context:} It \texttt{[event1]} \textbf{accused} \texttt{[/event1]} the company of deliberately slashing oil revenues by overproducing oil and \texttt{[event2]} \textbf{driving} \texttt{[/event2]} down prices, among other charges. \textit{Answer the question:} Does \texttt{[event1]} \textbf{accused} \texttt{[/event1]} happen after \texttt{[event2]} \textbf{driving} \texttt{[/event2]}? YES\\
         \hline
        \textbf{ \textit{QA\textsubscript{2}}}  & \textit{Given the context:} It \texttt{[event1]} \textbf{accused} \texttt{[/event1]} the company of deliberately slashing oil revenues by overproducing oil and \texttt{[event2]} \textbf{driving} \texttt{[/event2]} down prices, among other charges. \textit{Answer the questions:} Does \texttt{[event1]} \textbf{accused} \texttt{[/event1]} happen after \texttt{[event2]} \textbf{driving} \texttt{[/event2]}? YES Does \texttt{[event1]} \textbf{accused} \texttt{[/event1]} happen before \texttt{[event2]} \textbf{driving} \texttt{[/event2]}? NO Does \texttt{[event1]} \textbf{accused} \texttt{[/event1]} happen at the same time as \texttt{[event2]} \textbf{driving} \texttt{[/event2]}? NO\\
    \end{tabular}
    \caption{Prompt schema used in ICL. In QA\textsubscript{1} we report one of the questions only, but the schema is the same for the others. }
    \label{tab:prompts}
\end{table*}

\begin{table*}[htpb]
    \centering
    \renewcommand{\arraystretch}{1.6}
    \begin{tabular}{p{10cm}|c|c}
         \textbf{Context} & \textbf{Gold} & \textbf{Prediction}\\
         \toprule
         Evana Roth told CNN in August she \textbf{\textit{e1:}thought} her husband devised the plan after he was fired from his job in July. Her attorney, Lenard Leeds, said she had been unaware of the ruse before she \textbf{\textit{e2:}uncovered }the e-mail correspondence. & BEFORE & AFTER\\
         \hline
        The US embassy in Moscow has voiced concern and \textbf{\textit{e1:}asked} the Russian government for an explanation. A new Russian law \textbf{\textit{e2:}says} foreign-funded non-governmental groups (NGOs) linked to politics must register as "foreign agents" - a term which suggests spying.& AFTER & BEFORE\\
         
    \end{tabular}
    \caption{Challenging examples mispredicted by LLMs and RoBERTa-based model. The examples are taken from the MATRES corpus.}
    \label{tab:challenging_examples}
\end{table*}


\begin{table*}[!ht]
\centering
\renewcommand{\arraystretch}{1.2}
\begin{tabular}{p{3.2cm} |ccc|ccc|ccccc}
    \toprule
     \multicolumn{1}{c|}{\textbf{Models}} & \multicolumn{3}{c|}{\textbf{MATRES}}  & \multicolumn{3}{c|}{\textbf{TIMELINE}} & \multicolumn{5}{c}{\textbf{TB-Dense}} \\
     \toprule
     & Bef & Aft & Eq & Bef & Aft & Eq & Bef & Aft & Eq & Incl & Is\_Incl \\
     \toprule

      Mistral 7B & 71.4 & 1.7 & 1.9 & 64.0 & 2.4 & 5.2 & 5.7 & 0.0 & 3.3 & 3.4 & 0.0 \\
      Mixtral 8$\times$7B & 73.9 & 0.6 & 1.8 & 70.9 & 0.1 & 3.7 & 25.7 & 1.4 & 1.5 & 0.0 & 7.7 \\
      Llama2 7B & 71.9 & 6.4 & 2.1 & 70.1 & 25.7 & 0.0 & 0.2 & 30.1 & 0.3 & 9.5 & 0.0 \\
      Llama2 13B & 2.0 & 53.5 & 0.7 & 5.0 & 57.7 & 2.5 & 10.6 & 41.9 & 3.3 & 11.0 & 5.1 \\
      Llama2 70B & 76.8 & 33.6 & 0.0 & 74.8 & 24.1 & 0.0 & 0.3 & 31.7 & 3.6 & 0.0 & 0.0 \\
      GPT-3 & 67.2 & 26.9 & 0.0 & 68.4 & 32.9 & 0.0 & 0.0 & 3.1 & 2.6 & 5.0 & 1.1 \\
      GPT-3.5 & 72.4 & 36.8 & 0.0 & 70.3 & 30.0 & 9.5 & 36.1 & 0.7 & 0.0 & 0.0 & 0.0 \\
      Llama2 7B\textsubscript{Fine-tuned} & 87.2 & 77.9 & 0.0 & 81.9 & 81.2 & 0.0 & 63.7 & 34.6 & 0.0 & 0.0 & 0.0 \\
      Llama2 13B\textsubscript{Fine-tuned} & 88.6 & 82.1 & 0.0 & 68.3 & 51.8 & 0.0 & 65.8 & 51.9 & 0.0 & 0.0 & 0.0 \\
      \bottomrule
      RoBERTa & 91.8 & 86.2 & 0.0 & 89.8 & 87.4 & 8.9 & 88.6 & 86.7 & 0.0 & 56.6 & 59.2 \\
      \bottomrule
     
    \end{tabular}
    \caption{ F1-scores in percentage for each class of each dataset achieved using the best prompt settings.
        \texttt{(Bef=Before, Aft=After, Eq=Equal,  Incl=Includes, Is\_Incl=Is Included)}
    }
\label{tab:f1_scores_all}
\end{table*}

\end{document}